\documentclass[cameraready]{Interspeech}

\title{Predicting Cognitive Load from Speech and Interaction Dynamics in Dyadic Conversations}

\author[ orcid=0000-0001-7219-2150]{Tahiya}{Chowdhury}

\address{
    Department of Computer Science, Colby College, Waterville, Maine, USA
}

\email{tahiya.chowdhury@colby.edu}


\newcommand{\blue}[1]{\textcolor{black}{#1}}

\usepackage{comment}
\usepackage{booktabs}
\usepackage{multirow}

\begin{document}

\maketitle

\begin{abstract}
Estimating cognitive load from speech has largely been studied in controlled laboratory settings, with limited understanding of its reliability in natural collaborative conversations. We investigate whether speech and interaction dynamics predict perceived cognitive load during dyadic conversations. We analyze audio from \blue{53 dyads} performing nine collaborative tasks and extract static acoustic, dynamic, and interaction features to train a two-head Gated Recurrent Unit encoder to predict cognitive load scores. Results show conversational interaction provides useful signals for predicting cognitive load related to time pressure, mental work, effort, and task performance. Temporal demand is associated with turn-taking dynamics such as overlap and speaker switch, while mental demand is linked to imbalanced participation between speakers. These findings highlight the importance of task structure and conversational interaction for modeling cognitive load in natural collaborative settings.
\end{abstract}

\section{Introduction}
The recent rapid shift toward remote and hybrid work has intensified reliance on voice-mediated collaboration across geographically distributed teams. Prior research in safety-critical domains such as driving and construction shows that cognitive overload from voice communication during a task can degrade task performance, increase error rates, and compromise safety~\cite{miller2018voice, ranney2005effects, reagan2013using, abel2017cognitive}. In knowledge-work settings, remote meetings and collaborative tasks demand sustained attention and coordination, which can impair decision-making ability and well-being. Subjective instruments such as the NASA Task Load Index (NASA-TLX)~\cite{hart1988development} remain the gold standard for measuring perceived workload, yet they lack the fine-grained temporal resolution needed to be useful for real-time monitoring. As a possible solution, Speech-based vocal biomarkers have been used as non-invasive and ecologically valid indicators of stress and workload in naturalistic conversation for high stake domains such as aircraft and air traffic control~\cite{vukovic2021cognitive, yang2023cognitive, Gnerre2026, Diarra2025}, operation simulation~\cite{vukovic2019estimating}, and disaster response~\cite{khawaja2012analysis, vlachostergiou2020see}. 

While prior work has investigated cognitive load detection from speech, estimating cognitive load has been mostly formulated as a multi-class classification task. Early studies show that acoustic features such as pitch and energy could classify discrete workload levels above chance~\cite{Yin2007, Yin2008, Schuller2014}, an approach continued in follow-up works that employed binary vector representations~\cite{VanSegbroeck2014}, CNN-based architectures~\cite{Vukovic2021, yang2023cognitive}, and multimodal fusion methods~\cite{Xu2025, Liu2023}. Recent works have used self-reported NASA-TLX measures as ground truth for personalized load estimates from speech~\cite{Spang2023, vukovic2019estimating}. Across aviation, listening effort, and collaborative virtual environments, acoustic information such as fundamental frequency, loudness, and speech rate has been found informative for workload~\cite{Seeman2015, Boyer2018, Taptiklis2023}.

Despite these advances, several important gaps remain. First, most prior studies discretize cognitive load and treat its prediction as a classification problem, which can obscure meaningful continuous variation occurring during an interaction. Second, evaluation protocols often rely on random test splits, risking overly ambitious performance, yet lack generalization and stability in out-of-sample testing (unseen dyads). 
Third, most works investigated cognitive load from speech from data collected for a single task~\cite{Taptiklis2023, Boyer2018, VanSegbroeck2014}, which limits our understanding of cognitive load for dyads involving multiple tasks and task-specific influence on load.
Finally, while speech-based cognitive load prediction using individual acoustic features is well studied, comparatively fewer works investigate the contribution of temporal dynamics (temporal derivatives) and interaction features (turn-taking, overlap, speaker role switch), which are key characteristics in conversational settings. 

In this work, we address these gaps by investigating cognitive load prediction from speech in dyadic conversations. We model cognitive load as a regression task and evaluate performance using Concordance Correlation Coefficient (CCC), Pearson correlation, and RMSE following prior works~\cite{nguyen2024mental, hannula2008comparison}. We systematically compare the contribution of various feature sets, including static acoustic, temporal acoustic, and interaction features, to quantify the relative contribution of fine-grained information about acoustic, temporal dynamics, and conversational structure to model perceived cognitive load. We focused our evaluation on cross-dyad generalization for a realistic assessment of speech-based workload prediction in remote collaboration and conversational systems. We aim to answer the following questions in this work:
\begin{itemize}
    \item Can we model cognitive load during dyadic conversation as a regression task using speech samples?
    \item Does interaction and temporally dynamic acoustic features provide complementary signals that improve cognitive load prediction for dyadic conversations?
    \item To what extent does predictive signal reflect task-specific interaction pattern rather than cognitive load itself?
\end{itemize}

\section{Data}

In this work, we used AVCAffe~\cite{sarkar2023avcaffe}, an audiovisual dataset of remote collaboration and conversation recorded over a video-conferencing platform. Each conversation involves a dyad collaborating to complete up to nine tasks of varying levels of cognitive demand. We choose this dataset for several reasons: 1) the dataset is not collected from the internet, 2) it simulates real-world remote collaboration, 3) it provides per-task cognitive load annotation, which has been less explored, and 4) the tasks resemble remote collaboration where people need to collaborate together to complete multiple tasks of varying difficulty.

The 106 participants are recruited from 18 different countries of origin, with ages ranging between 18 and 57 years old \footnote{75\% from the age group of 21--30 years, with 52 males, 53 Females, and 1 non-binary. Audio files collected at 44.1 kHz frequency.} The complete dataset has a total of 950 per-participant recordings with self-
reported ground truth labels by each participant for affect (arousal, valence, dominance) and cognitive load attributes such as mental demand, temporal demand, effort, frustration, performance, and physical demand derived from the NASA Task Load Index. 

The nine tasks include open discussion (7.5 minutes), lighten mood with jokes (7.5 minutes), find differences in pictures (10 minutes), two map matching (2.5 minutes), Lost at Sea (10 minutes), two reading comprehensions (5 minutes each), and a multi-task. The tasks are chosen to induce varied levels of cognitive states at different stages of the conversation, reflecting a remote meeting setup where dyads are involved in open-ended discussion, complex problem solving, and decision making in collaboration.

\subsection{Data Pre-processing}
We extracted a single-channel audio stream per participant per task and resampled at 16 kHz. 
We then partition each task-level audio into contiguous, non-overlapping windows of $W=30$s, resulting in an average of $~11.6$ windows per task (min: 2, max: 27), which builds our speech dataset.

Because conversational tasks contain long pauses and unequal speaking time between speakers (especially in problem-solving tasks), some windows can be silent. To avoid acoustic feature extraction on such windows, we applied Silero VAD, a pre-trained voice activity detector~\cite{Silero_VAD}, to each window to estimate speech activity and compute measures such as total seconds of detected speech in the window, fraction of speech in the window to identify insufficient speech activity, and mask them for feature extraction. This acts as a timing proxy for calculating interaction dynamics-related features later. 

\section{Method}

\subsection{Features}
\label{Feature}

\textbf{Static Acoustic Features.} For any sample with speech activity, we use OpenSMILE~\cite{eyben2010opensmile} with its eGeMAPSv02 (extended Geneva Minimalistic Acoustic Parameter Set)~\cite{eyben2015geneva} \blue{feature set}, designed and widely used in voice research and affective computing. The feature set includes prosodic, spectral, cepstral, and voice-quality descriptors (e.g., fundamental frequency, loudness, spectral slopes, jitter, shimmer, etc. measures) resulting in 88 acoustic dimensions per window.

\textbf{Temporally Dynamic Acoustic Features.}
Prior work suggests that cognitive load varies over time based on task complexity~\cite{wirzberger2017embedded}. To encode short-term speech dynamics and capture within-task temporal change, we compute first-order differences over time as: $\Delta \mathbf{x}_t = \mathbf{x}_t - \mathbf{x}_{t-1}$ for each acoustic feature, with $\Delta$ being zero for the first window. This generates a second set of 88 features.

\textbf{Interaction Features.}
For each dyad-task, we also compute number of windows available for both speakers, fraction of windows where both members speak, fraction of windows where neither member speaks, fraction of windows where A only and B only speaks, mean speaking fractions for both A and B, difference between speech fraction (to indicate dominance), turn switches, turn switch rate, rate of switches between A-only and B-only. These features are intended to capture dyadic interaction using only timing information, independent of lexical content. We include them as an additional feature set to test the predictive value of interaction features.

\subsection{Cognitive Load Labels} Cognitive load scores are collected following the NASA Task Load Index protocol, a standard subjective workload instrument~\cite{hart1988development, hart2006nasa} on a scale of 0-21 across the six dimensions. Cognitive load scores are collected at the end of each task, which appear in the same order for each, and we use these as cognitive load labels. We use NASA-TLX scores per participant as participant-specific targets and train a two-head predictor (one head per participant).

\subsection{Regression Models}
We model cognitive load as a regression task. 
For each dyad-task instance, we align participants by window index and form paired sequences $(\mathbf{X}^A, \mathbf{X}^B)$ where $\mathbf{X}^A \in \mathbb{R}^{T \times F}$ and $\mathbf{X}^B \in \mathbb{R}^{T \times F}$, with appropriate padding as our input sample. The final dataset includes 475 input samples from 950 tasks.

\textbf{Temporal model: GRU.} We use a shared gated recurrent unit (GRU) encoder~\cite{cho2014properties} to model temporal speech dynamics in the samples, which has been empirically successful in speech signal modeling~\cite{chung2014empirical}. For each participant sequence, we apply mean pooling across time steps to obtain fixed-length embeddings. The two participant embeddings are concatenated and passed through a fully connected layer (128 hidden units) with ReLU activation and dropout (0.2). We train the model using a joint mean squared error (MSE) loss summed for both participants: $\mathcal{L} =
\mathrm{MSE}(\hat{y}_A, y_A) +
\mathrm{MSE}(\hat{y}_B, y_B)$
and two regression heads predict participant-level individual score (A and B) and dyad-level score by averaging individual scores.

\textbf{Baseline model: Random Forest.}
We use window-level features aggregated over the entire task duration. The resulting vectors are concatenated for both participants and used to train a Random Forest (RF) Regression model\footnote{no. of estimators = 300, minimum no. of leaf = 2}. Unlike the GRU, RF does not model temporal order and relies on aggregated acoustic summaries.

\subsection{Evaluation}
Considering our small sample size, to ensure generalization for out-of-sample testing (unseen dyads), we performed Leave-One-Dyad-Out (LODO) cross-validation. We chose this over commonly used k-fold cross-validation to avoid data leakage and get an unbiased estimate of our model performance. We standardize features \emph{within each fold} using training dyads only. The GRU model is trained for 25 epochs using Adam optimization with a learning rate of $10^{-3}$. Results are averaged across 10 random seeds (0-9) to assess performance stability.

\textbf{Metric.} We report performance using Concordance Correlation Coefficient (CCC) as the primary metric for its reliability as a measure of both correlation and agreement~\cite{o2017continuous, brady2016multi, toisoul2021estimation}, which is ideal for our cross-dyad generalization focus. We also include Pearson correlation (PCC) and Root Mean Squared Error (RMSE) to provide complementary information on linear association and absolute error magnitude, respectively, based on their usage in affective computing literature~\cite{eyben2012multitask, galvao2021predicting, han2017strength}. 

\begin{table}[t!]
  \caption{Cognitive load prediction performance using acoustic features across six workload dimensions with GRU model. Scores reported from 10-seed runs (mean $\pm$ std) to show stability. \blue{Bold indicates best performance.}}
  \label{tab:full_6load_results}
  \centering
  \setlength{\tabcolsep}{1pt}
  \begin{tabular}{l  l  c c c}
    \toprule
    \textbf{Dimension} & \textbf{Metric} & \textbf{Participant A} & \textbf{Participant B} & \textbf{Dyad Mean} \\
    \midrule
    
    \multirow{3}{*}{Temporal} 
    & CCC  & $0.34 \pm 0.03$ & $0.32 \pm 0.02$ & $\mathbf{0.42 \pm 0.03}$ \\
    & PCC  & $0.40 \pm 0.03$ & $0.37 \pm 0.02$ & $\mathbf{0.46 \pm 0.03}$ \\
    & RMSE & $6.26 \pm 0.10$ & $6.24 \pm 0.12$ & ${5.14 \pm 0.11}$\\
    \midrule
    
    \multirow{3}{*}{Mental} 
    & CCC  & $\mathbf{0.31 \pm 0.04}$ & $0.13 \pm 0.03$ & $0.22 \pm 0.03$ \\
    & PCC  & $\mathbf{0.36 \pm 0.03}$ & $0.16 \pm 0.04$ & $0.25 \pm 0.03$ \\
    & RMSE & $5.48 \pm 0.09$ & $6.36 \pm 0.30$ & $4.78 \pm 0.17$ \\
    \midrule

    \multirow{3}{*}{Effort} 
    & CCC  & $0.23 \pm 0.05$ & $0.11 \pm 0.04$ & $0.20 \pm 0.06$ \\
    & PCC   & $0.29 \pm 0.04$ & $0.14 \pm 0.04$ & $0.25 \pm 0.05$\\
    & RMSE & $5.38 \pm 0.10$ & $5.99 \pm 0.08$ & $4.49 \pm 0.06$\\
    \midrule

    \multirow{3}{*}{Performance} 
    & CCC  & $0.16 \pm 0.03$ & $0.12 \pm 0.03$ & $0.19 \pm 0.04$ \\
    & PCC  & $0.23 \pm 0.03$ & $0.17 \pm 0.04$ & $0.24 \pm 0.03$\\
    & RMSE & $5.58 \pm 0.09$ & $5.92 \pm 0.12$ & $4.83 \pm 0.10$\\
    \midrule

    \multirow{3}{*}{Frustration} 
    & CCC  & $0.08 \pm 0.02$ & $0.03 \pm 0.02$ & $0.10 \pm 0.03$ \\
    & PCC  & $0.10 \pm 0.02$ & $0.04 \pm 0.03$ & $0.12 \pm 0.03$\\
    & RMSE & $6.13 \pm 0.16$ & $5.54 \pm 0.15$ & $4.45 \pm 0.13$\\
    \midrule

    \multirow{3}{*}{Physical} 
    & CCC  & $0.05 \pm 0.03$ & $0.17 \pm 0.02$ & $0.09 \pm 0.02$ \\
    & PCC  & $0.05 \pm 0.03$ & $0.20 \pm 0.02$ & $0.10 \pm 0.02$ \\
    & RMSE & $4.49 \pm 0.07$ & $3.82 \pm 0.06$ & $3.00 \pm 0.05$\\
    \bottomrule
  \end{tabular}
\end{table}


\section{Results}

\textbf{1. Can we model cognitive load during dyadic conversations as a regression task using speech samples?}\\

To answer this question, we trained our baseline (RF) and GRU-based neural encoder model with static acoustic features to predict all six dimensions of task load. Note that $n$ models are trained on $n-1$ dyads, tested on hold-out dyads ($n=53$). We report performance for predicting both participant A and participant B's individual load score, along with the dyad-level mean score. We present the results in Table~\ref{tab:full_6load_results}.

For temporal demand, which measures the time pressure perceived by participants, prediction performance ($CCC_{dyad}$ = $0.42$, $PCC_{dyad}$ = $0.46$) indicates that speech contains modest predictive signals of temporal load during dyadic interaction. We also observe that individual temporal load is shared by the two participants for this dimension (\blue{individual $CCC$ = $0.32-0.34$)}, indicating temporal load during dyadic interaction may arise from interaction itself. 

For mental demand, which measures the perceived thinking and analytical load, the prediction result indicates a load-related signal in speech ($CCC_{A}$ = $0.31$, $PCC_{A}$ = $0.36$) is not sufficiently strong to capture dyad-level load. Unlike temporal demand, there is an asymmetry in individual-level prediction: Participant A achieves substantially higher $CCC$ and $PCC$ than Participant B, suggesting a stronger predictive signal in the first speaker's speech. We observe a similar pattern for effort and performance, the measures for effort and the perceived level of success in accomplishing a given task. We \blue{exclude} the remaining two dimensions \blue{from further analysis}, as their results indicate a null predictive signal.
Considering the stable performance (averaged over 10 runs) on unseen dyads, we conclude that speech contains a modest but generalizable workload signal for temporal and mental demand.
\\

\begin{table}[t!]
\caption{Dyad-level Mean CCC under leave-one-dyad-out cross validation using regression model variants.}
\centering
\small
\begin{tabular}{lcc}
\toprule
Model & Mental  & Temporal  \\
\midrule
Random Forest & $0.22$  $\pm$ $0.01$ & $0.33$ $\pm$ $0.01$ \\
GRU      & $0.23$ $\pm$ $0.02$ & $0.41$  $\pm$ 0.01\\
GRU (w/ attn) & $0.22  \pm 0.03$ & \textbf{0.43  $\pm$ $0.03$} \\
\bottomrule
\end{tabular}
\label{Tab: Models}
\end{table}

\begin{table}[t!]
\caption{Prediction performance under leave-one-dyad-out protocol to predict 4 load dimensions with feature set variants (A: Acoustic, I: Interaction, T: Temporal).}
\label{tab:feature_sets}
\centering
\begin{tabular}{l|cccc}
\toprule
Feature & Temporal & Mental & Effort & Performance \\
\midrule
Acoustic (A) & 0.42 & 0.22 & 0.20 & 0.19\\
\hline
Temporal (T) & 0.35 & 0.27 & 0.15 & 0.21\\
A + T & 0.40 & 0.25 & 0.29 & 0.21\\
\hline
Interaction (I) & \textbf{0.51} & 0.28 & 0.13 &  0.16\\
A + I & 0.46 & \textbf{0.32} & \textbf{0.34} &  \textbf{0.31}\\
\bottomrule
\end{tabular}

\end{table}

\textbf{Modeling Choice.} To justify our modeling choice, we compare regression performance using random forest as a baseline and two variants of GRU encoder: with and without attention. We hypothesized that the attention mechanism, due to its prior success in focusing on the salient features in a sequence, would better model the temporal structure and interaction within the sequence and provide better predictions. We present dyad-level predictions for mental and temporal demand from all three models in Table~\ref{Tab: Models}. We do not observe GRU to be advantageous over Random Forest for predicting mental demand because the mean $CCC$ remains largely similar. However, for temporal demand, we observe better performance when using GRU over RF, but not from using attention. This may be attributed to the small sample size of our dataset, which is insufficient to leverage the temporal modeling \blue{capacity} of attention. We use GRU without attention as our base model for the rest of the experiments.

To determine whether GRU significantly outperformed RF, we compared the difference in per-dyad level performance from random forest and GRU using the Wilcoxon signed-rank test, a non-parametric paired statistical test across all \blue{53 dyads}. The Holm-Bonferroni correction was applied to control the family-wise error rate, as we tested multiple workload dimensions. Surprisingly, the differences were not statistically significant (Temporal demand:
$p=0.21$, corrected $p$
= 0.41; Mental demand: $p=0.20$, corrected $p=0.41$) indicating that the GRU model does not significantly outperform the Random Forest baseline for predicting cognitive load in this dataset.
\\

\begin{figure}[t!]
  \centering
  \includegraphics[width=0.8\linewidth]{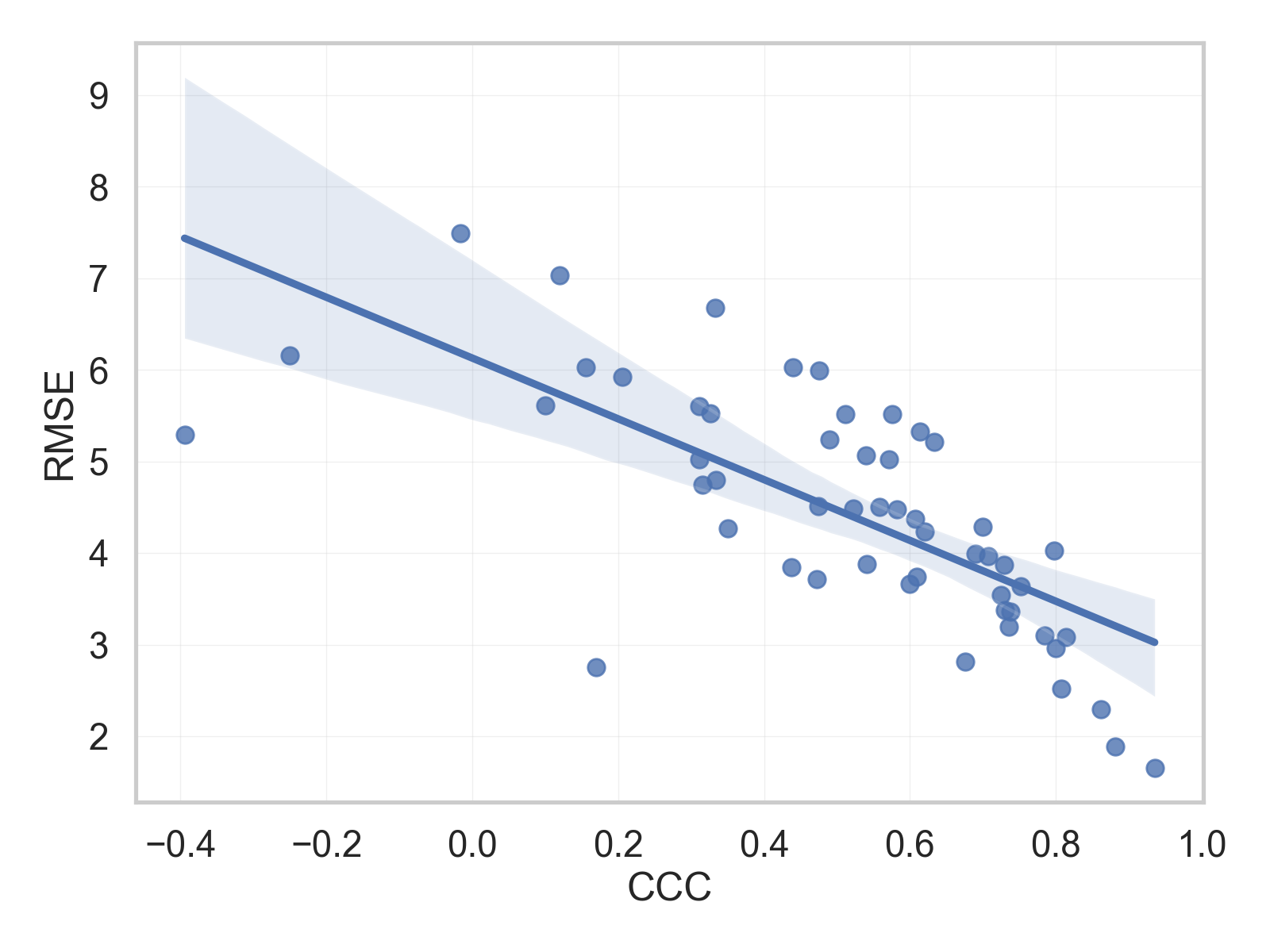}
  \caption{Per dyad CCC vs. RMSE. Note that CCC for some dyads is as high as $0.6-0.9$ indicating high variability across dyads.}
  \label{fig:per_dyad_ccc_vs_rmse}
  \vspace{-10pt}
\end{figure}

\textbf{2. Do dynamic acoustic and interaction features provide complementary signals to acoustic features that improve cognitive load prediction?}\\

To answer this, we assessed prediction performance for 4 workload dimensions (excluding frustration and physical demand) with different feature sets used for modeling. Specifically, we considered three primary sets (section~\ref{Feature}): static acoustic (A: 88 eGeMAPS features), temporally dynamic acoustic (T: 88 eGeMAPS feature delta over consecutive windows), and interaction features (I: 10 conversational interaction-related features based on speaking time, turn-taking, and coordination). We also created two composite feature sets (Acoustic + Temporal and Acoustic + Interaction, respectively) and used a GRU model for predicting cognitive load.

The $CCC$ scores are presented in Table~\ref{tab:feature_sets}. We observe that temporal features only or adding them to the acoustic do not improve performance generally for all four load dimensions. Only the temporal demand experienced a small increase. However, with 10 interaction features, $CCC$ increased $0.42 \rightarrow 0.51$ for temporal and interaction features and $0.22 \rightarrow 0.32$ when interaction features combined with acoustic, suggesting that the interaction-related features carry substantial predictive signal that can generalize across dyads. By combining acoustic with interaction-related features, all prediction performances for the four cognitive loads improved.

We conclude that temporal demand (with its higher $CCC$ ~0.5) is closely related to turn-taking and coordination during conversational interaction. Interaction features capture \blue{these} coordination dynamics, hence they generalize across dyads. However, these features may also reflect task structure rather than cognitive load alone, highlighting the importance of disentangling task-related effects from workload-related signals.\\

\blue{\textbf{3. To what extent does predictive signal reflect task-specific interaction pattern rather than cognitive load itself?}\\}

\textbf{Dyad Variability.} We observed from our earlier experiments that cognitive load prediction performance varied for different load dimensions. While dyad-level mean $CCC$ is $0.51$, which is moderate, it weakens the dyad-specific signal. $CCC$ for each of \blue{53 dyads} varies substantially, which indicates high heterogeneity across dyads (mode = $0.77$). This is salient in Figure~\ref{fig:per_dyad_ccc_vs_rmse}, where we plot per dyad CCC (x-axis) and RMSE (y-axis). \blue{For some dyads, CCC is as high as \blue{$0.6-0.9$}, indicating high correlation and agreement between predicted and ground truth value, while an inverse relationship ($-ve$) for at least two dyads.} As $CCC$ increases, RMSE decreases, showing models with better agreement also reduce error in absolute load value estimation. Overall, we concluded that while interaction features capture coordination dynamics that are helpful to generalize perceived workload prediction across dyads, variability across dyads remains substantial, which weakens the aggregated results. This variability may come from differences in speaking styles, expression (considering the geographically diverse dataset), and task completion strategies.

\begin{figure}[t!]
  \centering
  \includegraphics[width=0.99\linewidth]{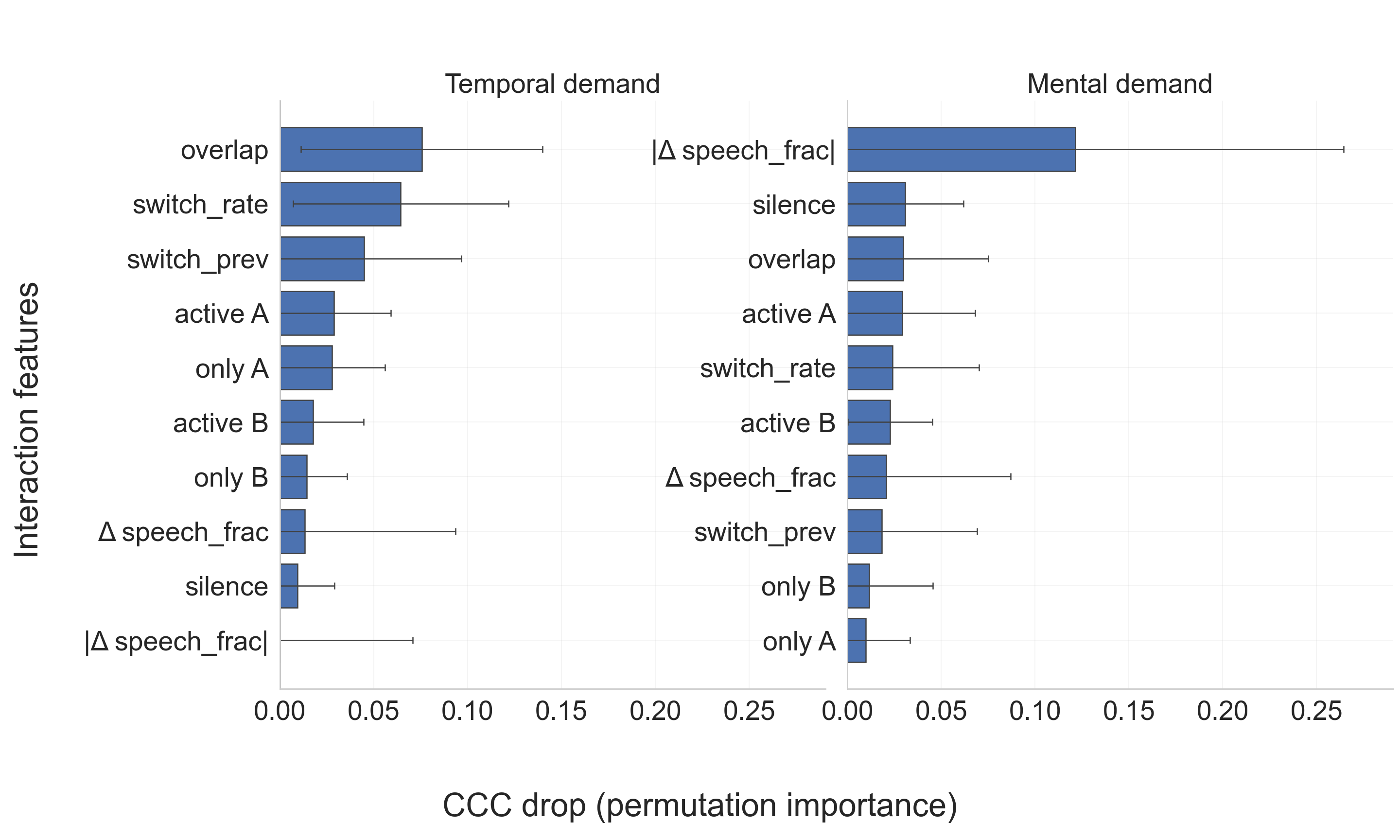}
  \caption{Permutation feature importance for interaction features in predicting temporal and mental demand. Bars indicate the mean CCC drop when each feature is permuted, with whiskers indicating the standard deviation across folds.}
  \label{fig:permulation_feat_imp}
  \vspace{-10pt}
\end{figure}

\textbf{Salient Features.} We also inspected how different interaction dynamics influence different cognitive load dimensions by permuting each interaction feature and tracking the drop in $CCC$. The results of this feature importance analysis (Figure~\ref{fig:permulation_feat_imp}) reveal that temporal demand is primarily associated with turn-taking dynamics and conversation pacing, such as overlap, speaker switching rate, and switching from the previous speaker, suggesting that perceived time pressure emerges from interaction coordination. Time constraints likely made participants interrupt and overlap with each other more, and cause more frequent turn exchanges, which is supported by prior works on conversational behavior~\cite{johnstone1995there, goldman1952individual}. In contrast, mental demand is most strongly predicted by imbalance in speaking time between participants, indicating that collaborative and problem-solving tasks requiring mental and analytical thinking may lead to asymmetric participation patterns during the conversation, where one speaker contributes disproportionately more than the other (conversational dominance). These findings suggest that perceived workload emerges from both task-level interaction patterns and dyad-level conversational behaviors, highlighting the need for further research to disentangle these interplaying factors when modeling cognitive load in collaborative settings.

\section{Limitations and Future Work}

The work presented here is based on a relatively small dataset (\blue{53} dyads and 475 task samples), which limits the ability of sequence models like GRU with attention to fully leverage temporal conversational dynamics. Our results here are based solely on speech dynamics and conversational coordination, calculated from voice activity. Including richer multimodal signals, such as lexical, gaze, facial expression, and body movement, can further improve generalizability and provide nuanced insights into conversational behavior. Furthermore, workload measures are collected at the task level, so more fine-grained temporal modeling to capture the time-evolving nature of conversational dynamics during a task is left as a future direction. By understanding these dynamics, we will be able to better design speech technologies to support the way people speak and collaborate together.

\section{Acknowledgments}

We thank the AIIM lab, Queen's University, Canada, for collecting, preparing, and making this dataset publicly available. We also thank the reviewers for their helpful comments to improve this work. This work was supported by the Henry Luce Foundation.

\section{Generative AI Use Disclosure}

Generative AI (GPT 5.2) was used for LaTex table and figure formatting.


\bibliographystyle{IEEEtran}
\bibliography{reference}

\end{document}